\documentclass[journal]{IEEEtran}
\usepackage{amsmath}
\usepackage{graphicx}
\usepackage[hidelinks]{hyperref}
\usepackage{amsthm}
\usepackage{multirow}
\usepackage{booktabs}
\usepackage{algorithm}
\usepackage{algorithmic}
\usepackage{color}
\usepackage{amsfonts}
\usepackage{ulem}
\usepackage{bm}

\ifCLASSOPTIONcompsoc
  \usepackage[nocompress]{cite}
\else
  \usepackage{cite}
\fi
\ifCLASSINFOpdf
\else
\fi
\begin{document}
%
\title{Graph Domain Adaptation: A Generative View}
%
%
%
%

\author{Ruichu Cai*,~\IEEEmembership{Member,~IEEE,}
        Fengzhu Wu,
        Zijian Li, Pengfei Wei, Lingling Yi, Kun Zhang
\IEEEcompsocitemizethanks{\IEEEcompsocthanksitem Ruichu Cai is with the School of Computing, Guangdong University of Technology and and Guangdong Provincial Key Laboratory of Public Finance and Taxation with Big Data Application, Guangzhou China, 510006.
E-mail: cairuichu@gmail.com
\IEEEcompsocthanksitem Fengzhu Wu is with the School of Computing, Guangdong University of Technology, Guangzhou China, 510006.
E-mail: fzwu97@gmail.com
\IEEEcompsocthanksitem Zijian Li is with the School of Computing, Guangdong University of Technology, Guangzhou China, 510006.
E-mail: leizigin@gmail.com
\IEEEcompsocthanksitem Pengfei Wei is with National University of Singapore. E-mail: wpf89928@gmail.com
\IEEEcompsocthanksitem Lingling Yi is with Tencent Technology (SZ) Co., Ltd. E-mail: chrisyi@tencent.com
\IEEEcompsocthanksitem Kun Zhang is with the Department of Philosophy, Carnegie Mellon University, Pittsburgh, PA 15213 USA. E-mail: kunz1@cmu.edu}
\thanks{Manuscript received XX; revised XX; accepted XX. Date of publication XX XX, 2019; date of current version XX XX, 2019. Ruichu Cai and Zijian Li was supported in part by Natural Science Foundation of China (61876043, 61976052), Science and Technology Planning Project of Guangzhou (201902010058) and Guangdong Provincial Science and Technology Innovation Strategy Fund (2019B121203012).  (\emph{Corresponding author: Ruichu Cai.})
}
}

%
%

\markboth{}%
{Ruichu Cai \MakeLowercase{\textit{et al.}}: Graph Domain Adaptation: A Generative View}
%



\IEEEtitleabstractindextext{%
\begin{abstract}
Recent years have witnessed tremendous interest in deep learning on graph-structured data. Due to the high cost of collecting labeled graph-structured data, domain adaptation is important to supervised graph learning tasks with limited samples. However, current graph domain adaptation methods are generally adopted from traditional domain adaptation tasks, and the properties of graph-structured data are not well utilized. 
\textcolor{black}{For example, the observed social networks on different platforms are controlled not only by the different crowd or communities but also by the domain-specific policies and the background noise.}
Based on these properties in graph-structured data, we first assume that the graph-structured data generation process is controlled by three independent types of latent variables, i.e., the semantic latent variables, the domain latent variables, and the random latent variables. Based on this assumption, we propose a disentanglement-based unsupervised domain adaptation method for the graph-structured data, which applies variational graph auto-encoders to recover these latent variables and disentangles them via three supervised learning modules. Extensive experimental results on two real-world datasets in the graph classification task reveal that our method not only significantly outperforms the traditional domain adaptation methods and the disentangled-based domain adaptation methods but also outperforms the state-of-the-art graph domain adaptation algorithms.
\end{abstract}

}

\maketitle
\IEEEdisplaynontitleabstractindextext
\IEEEpeerreviewmaketitle

\section{Introduction}\label{sec:introduction}
Though deep learning on graph-structured data has achieved great success, similar to other deep learning methods, it heavily depends on labeled data. However, the high cost of collecting and labeling graph-structured data in real-world applications is unacceptable. Fortunately, unsupervised domain adaptation can figure out the aforementioned problem. Therefore, unsupervised domain adaptation on graph-structured data is important to various graph learning tasks. One example is the influence prediction on social networks with different communities \cite{7463496,chen2020mining} \footnote{A community is a group of users with the same interest.}, which aims to predict whether a given user will retweet. 
\begin{figure*}[htb]
	\centering
	\includegraphics[scale=0.80]{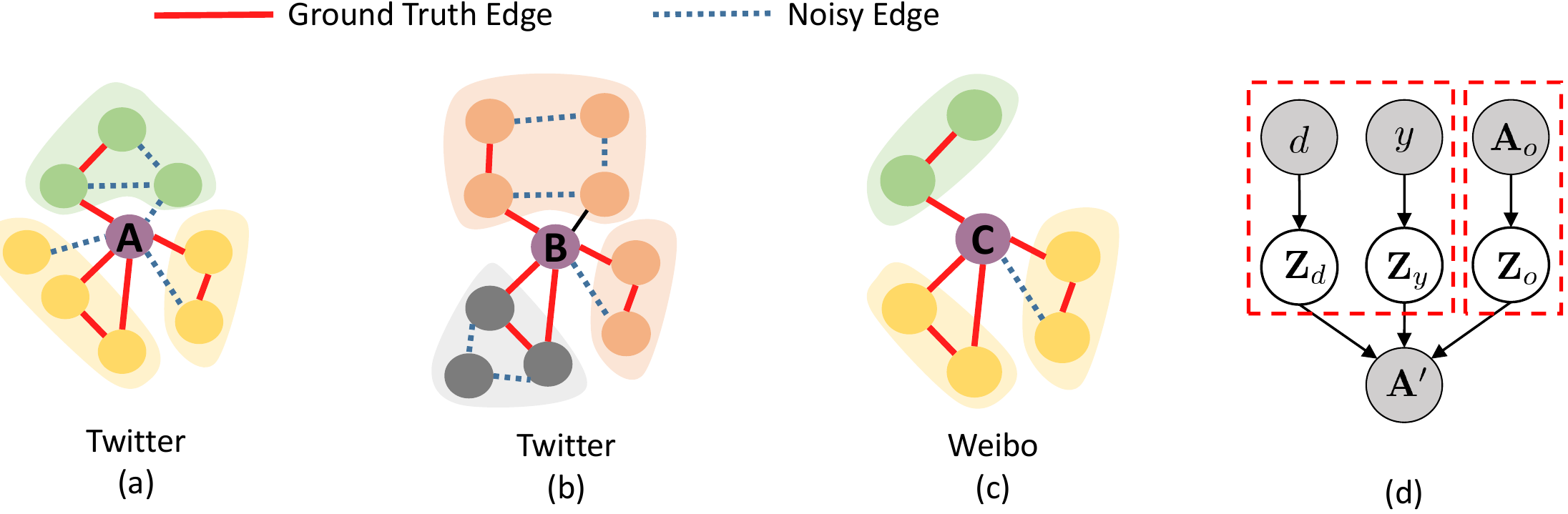}
	\caption{(a)-(c) Three observed graph structures from Twitter and Weibo. Each node denotes a user. The purple nodes denote the users who will retweet since they share the same structures. The irregular blocks in different colors denote different communities. Communities from different domains are not identical. The blue dashed lines and the red lines denote the noisy edges and the true edges, respectively.
	(d) The causal model of the graph-structured data generation process, which is controlled by three independent types of latent variables $\bm{Z}_d$, $\bm{Z}_y$ and $\bm{Z}_o$. Here, $d$, $y$, and $\bm{A}_{o}$ denote the domain information, semantic information, and the uncertainty of graph data. \textit{(best view in color)}}
	\label{fig:motivation}
\end{figure*}

Numerous unsupervised domain adaptation works have been designed for images data \cite{wang2019transferable}, texts data \cite{lin2018neural,HAO2021127} and time series data \cite{Cai_Chen_Li_Chen_Zhang_Ye_Li_Yang_Zhang_2021}, while only a few methods have been proposed for graph data. For instance, \cite{zhang2019dane} achieves domain-adaptive network embedding with the help of the shared graph convolutional network and the Generative Adversarial Networks (GAN) based adversarial regularization \cite{goodfellow2014generative}. \cite{wu2020unsupervised} applies the attention mechanism to integrate global and local consistency and extract cross-domain node embedding by deceiving the domain discriminator with the help of Gradient Reversal Layer (GRL) \cite{ganin2015unsupervised}. As summary, these adversarial methods are generally adopted from traditional domain adaptation tasks. {and the properties of graph-structured data are not explicitly utilized.} 


\textcolor{black}{Different from traditional data, the graph data is featured with the high uncertainty of the generation process and the high complexity of the domain-specific structures. We take the social network in Figure \ref{fig:motivation} (d) as an example. The observed social networks are not only controlled by the latent semantic factors (e.g., the shared interests among the nodes), but also by the domain-specific communities (e.g., raised by the domain-specified network policies) and the uncertainty (e.g., two nodes with the same interests may not know each other or the zombie fans in the social networks). The ignorance of these important properties for graph-structured data hinders the performance of the graph domain adaptation model. }


\textcolor{black}{In order to learn a graph representation that is robust across domains, two kinds of important information should be included. The first is the domain-invariant structure/topological information and the second is the domain-invariant node information. However, it is not a trivial task to extract this domain-invariant information at the same time. The main difficulty of extracting the domain-invariant structure information from the observed graph-structured data is the uncertainty. Due to the high uncertainty of observed graph data, even for the users surrounded by the same communities, the observed graph structures between different platforms/domains are totally different, e.g., the cases are given in Figure \ref{fig:motivation} (a) and (c). And the main obstacle of extracting the domain-invariant node information the different communities from different platforms/domains, which is usually led by the domain-specified network policies. Give a detailed example shown in Figure \ref{fig:motivation} (a) and (b), users with a similar observed graph structure, can be surrounded by different social communities. In this case, since the distribution and the types of communities vary across domains, it is hard to extract the domain-invariant node representation solely using the conventional domain adaptation restrictions like gradient reversal layer and MMD.}


Therefore, we propose a Disentanglement-based Graph Domain Adaptation Model (\textbf{DGDA} in short), which is motivated by the graph data generation process as shown in Figure \ref{fig:motivation} (d). In detail, we assume that the graph data generation process is controlled by three independent types of latent variables, i.e., the semantic latent variables $\bm{Z}_y$, the domain latent variables $\bm{Z}_d$, and the random latent variables $\bm{Z}_o$. These three groups of latent variables are encoded by the semantic information $y$, the domain information $d$, and the uncertainty $\bm{A}_o$, respectively. By reconstructing and disentangling these three latent variables, we can easily classify the label of different graphs based on $\bm{Z}_y$. Technically, we employ a variational graph auto-encoders to reconstruct these latent variables. We disentangle the semantic and domain latent variables by employing a label classifier and a domain classifier. We further disentangle the random latent variables by reconstructing the uncertainty of the graph data. Furthermore, we apply the latent variables regularization on these three types of latent variables for better disentanglement. The extensive experimental studies demonstrate that our method outperforms state-of-the-art unsupervised domain adaptation methods on graph data.

\textcolor{black}{The rest of the article is organized as follows. Section \ref{related_works} reviews existing studies on graph representation learning, domain adaptation, and graph domain adaptation. Section \ref{model} provides the problem definition on graph domain adaptation and the disentanglement-based graph domain adaptation model. Section \ref{experiment} presents the experiment results and analysis on two real-world datasets. Section \ref{conclusion} concludes the article with future work discussion.}

\section{Related work}\label{related_works}
\textcolor{black}{Our work is closely related to graph representation learning and domain adaptation, so we first review the existing techniques on graph representation learning and domain adaptation, and then we give an introduction about domain adaptation on graph-structured data.}
\subsection{Graph Representation Learning} Representation learning\cite{bengio2013representation} has been one of the hotspots of deep learning research fields. Graph representation learning aims to map edges and nodes of a graph into a low-dimensional vector space with original graph structures and properties being well preserved. Current methods can be categorized as inductive methods and transductive methods. Inductive methods \cite{kipf2016variational,wang2016structural,hamilton2017inductive,ying2018graph} learn functions that take the graph topology structure and edge or node features as input and output their representation vectors. Transductive methods \cite{tang2015line,dong2017metapath2vec,qiu2018network,kipf2016semi,velivckovic2017graph} optimize the representation vectors directly. Though these methods do not provide a representation of the entire graph, these methods are enough for our user-oriented prediction tasks. 

Recently, there have been several attempts to learn latent representations of sub-structures for graphs via kernel technique\cite{yanardag2015deep} or pooling operation. Global pooling methods aggregate the node representations either via simple readout functions such as averaging the node embeddings\cite{xu2018powerful} or more complex set operations\cite{vinyals2015order, zhang2018end}. On another line, hierarchical pooling methods
\cite{defferrard2016convolutional,ying2018hierarchical,ma2019graph,lee2019self,khasahmadi2020memory,huang2019attpool,gao2019graph} coarsen the node representations over the network’s layers\cite{mesquita2020rethinking}, and finally achieve the entire graph representation. 
However, these methods mainly focus on learning graph representation from a single domain and ignores the transferable sub-structures in the graph. 

\subsection{Domain Adaptation} Domain adaptation studies how to learn a model that is transferable on different but related domains. Previous studies mainly focus on learning domain invariant representation via neural networks, so that deep learning models trained on a labeled source domain can be transferred to a target domain with few or no labeled samples. They can be classified into statistic-based methods and adversarial methods. Statistic-based methods \cite{tzeng2014deep,long2015learning,Cai_Chen_Li_Chen_Zhang_Ye_Li_Yang_Zhang_2021} utilize maximum mean discrepancy (MMD) to achieve the domain alignment.  Inspired by GAN \cite{goodfellow2014generative}, adversarial domain adaptation methods \cite{ganin2015unsupervised,cai2019learning,zhang2019bridging,HAO2021127,shui2021aggregating,li2021causal} employ a domain adversarial layer to minimize the domain discrepancy, where a feature extractor and a domain classifier compete against each other. These methods mainly focus on grid data like images. In this paper, we focus on graph-structured data, which is more complicated and challenging than grid data.

\subsection{Graph Domain Adaptation}Recently, few methods have been proposed to address the graph domain adaption tasks. \cite{zhang2019dane} achieves domain adaptive network embedding via shared weight graph convolutional network and adversarial learning regularization. \cite{wu2020unsupervised} applies an attention mechanism to integrate global and local consistency and extract cross-domain node embedding with the help of Gradient Reversal Layer (GRL) \cite{ganin2015unsupervised}. \cite{shen2020adversarial} utilizes two feature extractors to preserve attributed affinity and topological proximity between nodes. 

However, these works mainly focus on the node classification, by simply replacing the feature extractor with the graph convolutional network, without taking the property of the graph-structured data into account. Recently, Elif et al \cite{pilanci2020domain,vural2019domain} handle graph domain adaptation via learning aligned graph bases. In this paper, we not only focus on the challenging graph classification task but also well utilize the property of graph-structured data via the generation process of graph-structured data.

\textcolor{black}{Please note that our work is related to but different from DIVA \cite{ilse2020diva}. The similarity is that both our method and DIVA aim to reconstruct three types of latent variables. The difference is that DIVA only reconstructs the residual variations in an unsupervised way,  while our method obtains the random latent variables by explicitly reconstructing three latent variables explicitly with the help of the supervised information as well as the synthetic random noise.} 

\begin{figure*}[htb]
	\centering
	\includegraphics[scale=0.55]{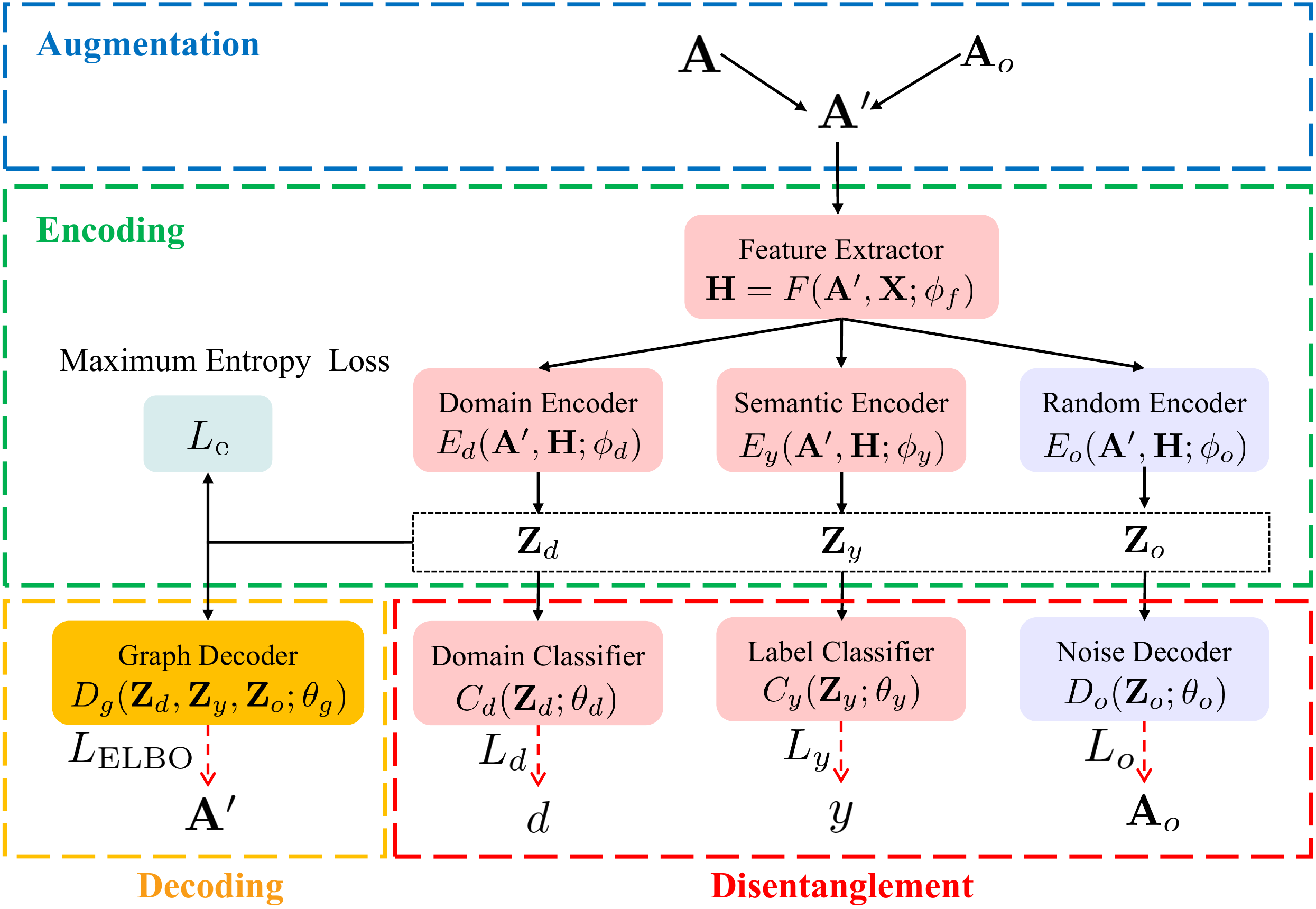}
	\caption{The framework of DGDA. 
    First, in the augmentation block, augmented samples are generated.
    Second, in the encoding block, a VGAE is applied to recover $\bm{Z}_d$, $\bm{Z}_y$, $\bm{Z}_o$.
    Third, in the decoding block, a graph decoder $D_g$ integrates information from all three latent variables to reconstruct $\bm{A}'$.
    Final, in the disentanglement block, three types of reconstruction are used to disentangle the latent variables. $C_y$ and $C_d$ are the classifiers for the label $y$ and the domain $d$, respectively. $D_o$ is used to reconstruct $\bm{A}_o$. $L_e$ is the latent variables regularization. \textit{(best view in color)}}
    \label{fig:model}
\end{figure*}

\section{Disentanglement based Graph Domain Adaptation Model} \label{model}

\newcommand{\tabincell}[2]{\begin{tabular}{@{}#1@{}}#2\end{tabular}}  
\begin{table}
	\centering
	\caption{Notation table.}
	\begin{tabular}{c|c}
		\hline
		\small{Symbols}  & Definitions and Descriptions \\
		\hline
		$G_{S}, n_{S}$ & The source domain dataset and its size.  \\
		\hline
		$G_{T}, n_{T}$ & The target domain dataset and its size.  \\
		\hline
		$g=\{\bm{A}, \bm{X}\}$ & \tabincell{c}{A graph sample in $G_{S}$ or $G_{T}$, \\ its adjacency matrix, and its node features.} \\
		\hline
		$d$ & The domain label of $g$. \\
		\hline
		$y$ & The graph label of $g$. \\
		\hline
		$N$ & The number of nodes in $g$. \\
		\hline
		$K_{\bm{X}}$ & The dimension of node features. \\
		\hline
		$\bm{Z}_{d}, E_{d}$ & The semantic latent variables and its encoder. \\
		\hline
		$\bm{Z}_{y}, E_{y}$ & The domain latent variables and its encoder. \\
		\hline
		$\bm{Z}_{o}, E_{o}$ & The random latent variables and its encoder. \\
		\hline
		$F$ & The node feature extractor. \\
		\hline
		$\bm{H}$ & The node representation outputs by $F$. \\
		\hline
		$\bm{\mu}$ & The mean of latent variables. \\
		\hline
		$\bm{\sigma}$ & The standard deviation of latent variables. \\
		\hline
		$C_{d}$ & The domain disentanglement module. \\
		\hline
		$C_{y}$ & The label disentanglement module. \\
		\hline
		$D_{o}$ & The noise reconstruction module. \\
		\hline
		$D_{g}$ & The graph reconstruction module. \\
		\hline
		$\bm{A}'$ & The augmented adjacency matrix of $g$. \\
		\hline
		$\bm{A}_{o}$ & The randomly generated noise matrix. \\
		\hline
		$\bm{D}$ & The diagonal degree matrix of $\bm{A}'$. \\
		\hline
		$\hat{\bm{A}}$ & The symmetric normalized $\bm{A}'$. \\
		\hline
		$\bm{W}, \bm{b}$ & Weights and biases in neural networks. \\
		\hline
		$\phi_{f}, \phi_{d}, \phi_{y}, \phi_{o}$ & The parameters in encoding block. \\
		\hline
		$\theta_{d}, \theta_{y}, \theta_{o}$ & The parameters in disentanglement block. \\
		\hline
		$\theta_{g}$ & The parameters of graph decoder $D_{g}$. \\
		\hline
		$\oplus$ & The concatenation operator of any two vectors. \\
		
		\hline
	\end{tabular}
	\label{tab:notation}
\end{table}

In this paper, we focus on the unsupervised domain adaptation problem on graph-structured data, which aims to use the labeled samples $G_{S}=\left\{{g}_{l}^{S}, y_{l}^{S}\right\}_{l=1}^{n_{S}}$ on source domain to classify the unlabeled samples $G_{T}=\left\{{g}_{l}^{T}\right\}_{l=1}^{n_{T}}$ on target domain, where $g$ and $y$ denote the graph sample and its label. Herein, $n_{S}$ and $n_{T}$ denote the number of samples in source and target domain, respectively. Each graph sample $g$ with $N$ nodes contains an adjacent matrix $\bm{A} \in \mathbb{R}^{N \times N}$ and a node feature matrix $\bm{X} \in \mathbb{R}^{N \times K_{\bm{X}}}$, i.e., $g=\{\bm{A}, \bm{X}\}$. The goal of this paper is to model the generation process of graph-structured data and design an effective graph domain adaptation framework.  The mathematical notations used in this paper are summarized in
Table \ref{tab:notation}.

The main motivation of our model is to address the high uncertainty of graph data and extract the semantic-related information behind the graphs. 
We start from the graph-structured data generation process as shown in Figure\ref{fig:motivation} (d). Given a graph sample, it assumed to be generated from three independent types of latent variables, i.e., $\bm{Z}_d$ encoding the domain information, $\bm{Z}_y$ encoding the semantic information, and $\bm{Z}_o$ encoding the uncertainty information. And $d$, $y$, and $\bm{A}_o$ respectively denote the domain information, the semantic information, and the uncertainty of graph data. We use $\bm{Z}_d \in \mathbb{R}^{N \times K_d}$ , $\bm{Z}_y \in \mathbb{R}^{N \times K_y}$ and $\bm{Z}_o \in \mathbb{R}^{N \times K_o}$ to denote domain latent variables, semantic latent variables and random latent variables. 
Considering that the domain information varies with domains, we induce that the semantic latent variables play an important role in extracting the domain-invariant representation. We further assume that these latent variables are independent.

Based on the aforementioned graph-structured data generative mechanism, we design a graph domain adaptation framework. The model architecture is shown in Figure \ref{fig:model}. We first extract features via a feature extractor $F(\cdot;\phi_f)$, and then use a VAE-like architecture to reconstruct three independent latent variables $\bm{Z}_d$, $\bm{Z}_y$ and $\bm{Z}_o$. 
However, unlike the vanilla VAE, we further design a disentanglement block as shown in the red dashed box in Figure \ref{fig:model}. In this architecture, three learning modules are placed under $\bm{Z}_d$, $\bm{Z}_y$, and $\bm{Z}_o$ respectively. For the domain disentanglement module $C_d(\bm{Z}_d;\theta_d)$, it aims to extract all the domain information into $\bm{Z}_d$ and exclude other information. For the label disentanglement module $C_y(\bm{Z}_y;\theta_y)$, it aims to extract all the semantic information into $\bm{Z}_y$ and exclude other information. For the noise variables reconstruction module $D_o(\bm{Z}_d;\theta_o)$, it aims to model the uncertainty of graph-structured data from $\bm{Z}_o$ and exclude other information. As a result, we obtain domain-invariant label information, which is also robust to uncertainty.
We will give a detailed description of our model in the following section.

\subsection{Uncertainty Modeling}
Real-world data is usually noisy, which leads to the degeneration of model performance. Data augmentation can mitigate this problem to a certain extent, but the unseen manipulations still can not be handle because it is impossible to see all different manipulations at the training stage. Motivated by \cite{zhang2020causal}, we apply a random manipulation-based data augmentation on graph-structured data. We explicitly model the distribution of noises via random latent variables $\bm{Z}_o$, which controls the uncertainty of the graph-structured data. In order to generate the augmented graph-structured data, we perform a valid augmentation of the original data, which means we further assume that this augmentation only controls the structure of $\bm{A}$ and is independent of the graph label $y$ and the domain label $d$. There are a variety of methods to generate augmented graph data. In this paper, we adopt a strategy similar to DropEdge\cite{rong2019dropedge}. For each graph sample, we first generate a noise matrix $\bm{A}_o$ by randomly dropping and adding edges in the original adjacency matrix $\bm{A}$. The details of $a_{ij} \in \bm{A}_o$ as shown in Equation(\ref{equ:noise_elem}). 
\begin{equation}\label{equ:noise_elem}
    a_{ij}=\left\{
    \begin{aligned}
        0& \quad\quad  \text{Constant between node $i$ and node $j$}.\\
        1& \quad\quad \text{Add a edge between node $i$ and node $j$}. \\
        -1& \quad\quad \text{Drop a edge between node $i$ and node $j$}.
    \end{aligned}
    \right.
\end{equation}
Specifically, the data augmentation process is formulated as follows:
\begin{equation}\label{equ:noise_gen}
    \begin{aligned}
        \bm{A}' & = \bm{M}_\text{drop} \odot \bm{A} + \bm{M}_\text{add}, \\
        {m}^\text{drop}_{ij} & \sim \text{Bernoulli}(p_\text{drop}), \\
        {m}^\text{add}_{ij} & \sim \text{Bernoulli}(p_\text{edge} \cdot p_\text{add}), \\
        \bm{A}_o & = \bm{A}' - \bm{A}, \\
    \end{aligned}
\end{equation}
where $p_\text{drop}$, $p_\text{add}$, and $p_\text{edge}$ refer to the dropping edge rate, adding edge rate, and the sparsity of $\bm{A}$, respectively. $\bm{M}_\text{drop}$ and $\bm{M}_\text{add}$ denote the mask matrix with the same shape as $\bm{A}$. 
${m}^\text{drop}_{ij}$ and ${m}^\text{add}_{ij}$ denote the elements in two mask matrix, and they are sampled from the Bernoulli distribution independently. 

Based on the aforementioned graph-structured data generative mechanism, we design a graph domain adaptation framework by first recovering the three types of latent variables via a VAE, and then disentangling them. 

\subsection{Latent Variables Reconstruction}

The encoding and decoding modules in our framework are inspired by Variational Graph Auto-Encoders (VGAE) \cite{kipf2016variational}, a framework for unsupervised learning on graph-structured data based on the VAE. In VGAE, $q_{\phi}(\bm{Z}|\bm{A}, \bm{X})$ denotes the encoder with respect to the parameter $\phi$ to approximate the intractable true posterior distribution $p(\bm{Z}|\bm{A}, \bm{X})$, while $P_{\theta_r}(\bm{A}|\bm{Z})$ denotes the decoder with respect to the parameters $\theta_r$. $P(\bm{Z})$ is the prior distribution. The variational lower bound of the marginal likelihood is given as follow:
\begin{equation}\label{equ:elbo}
\begin{split}
	\mathcal{L}_{\mathrm{ELBO}}(\phi , \theta_r)= \mathbb{E}_{q_{\phi}(\bm{Z}|\bm{A}, \bm{X})}[P_{\theta_r}(\bm{A}|\bm{Z})] \\
    - D_{KL}(q_{\phi}(\bm{Z}|\bm{A}, \bm{X}) \| P(\bm{Z})), 
\end{split}
\end{equation} where $D_{KL}(q(\cdot) \| P(\cdot))$ is the Kullback-Leibler divergence between distribution $q(\cdot)$ and $P(\cdot)$.

Unlike vanilla VGAE, we further decompose the latent variables $\bm{Z}$ into $\bm{Z}_{d}$, $\bm{Z}_{y}$ and $\bm{Z}_o$. As a result, Equation (\ref{equ:elbo}) can be derived as:
\begin{equation}
	\label{elbo_dyo}
	\begin{aligned}
    		& \mathcal{L}_{\mathrm{ELBO}}\left(\phi_{f, d, y, o}, \theta_{g}\right)=\\
	& \mathbb{E}_{q_{\phi_{d, y, o}}\left(\bm{Z}_{d}, \bm{Z}_{y}, \bm{Z}_{o} | \bm{H}\right)}\left[\log P_{\theta_{g}}\left(\bm{A}' | \bm{Z}_{d}, \bm{Z}_{y}, \bm{Z}_{o}\right)\right] \\
    &- \sum_{k \in (d, y, o)} D_{K L}\left(q_{\phi_{k}}\left(\bm{Z}_{k} | \bm{A}', \bm{X}\right) \| P\left(\bm{Z}_{k}\right)\right), 
	\end{aligned}
\end{equation},
where $\bm{H}=F\left(\bm{A}',\bm{X};\phi_f\right)$. $F$ is the feature extractor that takes the node features and the adjacency matrix as input, and outputs the node representation 
$\bm{H} \in \mathbb{R} ^ {N \times K_{\bm{H}}}$. 

We further assume that the prior distributions on all the latent variables are normal distributions, i.e., $P(\bm{Z}_{d}), P(\bm{Z}_{y}), P(\bm{Z}_{o}) \sim \mathcal{N}(\bm{0}, \bm{I})$. Similar to VGAE, by applying a reparameterization trick, we use three Graph Convolutional Networks (GCNs) \cite{kipf2016semi} based encoders $E_{d}$, $E_{y}$ and $E_o$ as the approximator of $q$, to encode the node representation $\bm{H}$ into $\bm{Z}_{d}$, $\bm{Z}_{y}$ and $\bm{Z}_{o}$, respectively. Finally, we aggregate all the latent variables and employ an inner product decoder that maps each pair of node representations to a binary indicator of edge existence in $\bm{A}'$. 

\subsubsection{Inference Model.}
We first extract the node representation via the GCNs. The GCNs comprises multiple stacked graph convolutional layers to extract features from multi-order neighbors. We take the node features $\bm{X}$ and the augmented adjacency matrix $\bm{A}'$ as the input of the GCNs, where each GCN layer can be formulated as follow:
\begin{equation}
\begin{split}
    \text{GCN}(\bm{A}', \bm{X})&=ReLU(\hat{\bm{A}} \bm{X} \bm{W}),\\ 
    \hat{\bm{A}}&=\tilde{\bm{D}}^{-\frac{1}{2}}\tilde{\bm{A}} \tilde{\bm{D}}^{-\frac{1}{2}},
\end{split}
\end{equation}
where $\tilde{\bm{A}}=\bm{A}'+\bm{I}_m$ is the augmented adjacency matrix with self-loop, $\tilde{\bm{D}}_{ii}=\text{diag}(\sum_{j} {\tilde{\bm{A}}_{ij}})$ is the diagonal degree matrix, $\bm{W}$ is the trainable parameters and $ReLU$ is the Rectified Linear Unit (ReLU) activation function. Specifically, in DGDA, we use a two-layer GCN as the feature extractor. For convenience, we let $\bm{H}=F(\bm{A}',\bm{X};\phi_f)$ be the aforementioned process, $\phi_f$ are the trainable parameters. 

In order to obtain three types of latent variables which are shown in Figure \ref{fig:motivation} (d), we further devise three encoders. These encoders are used to encode $\boldsymbol{\mu}$ and $log\boldsymbol{\sigma}$ of $\bm{Z}_{d}$, $\bm{Z}_{y}$ and $\bm{Z}_{o}$, respectively. For the type $k \in \{d, y, o\}$ latent variables, the encoding model is formulated as follows:
\begin{equation}
\begin{split}
    q(\bm{Z}_k |\bm{A}', \bm{H})=\prod_{i=1}^{N} q\left(\bm{z}_{ki} |\bm{A}', \bm{H}\right), \\ 
    \quad q\left(\bm{z}_{ki} |\bm{A}', \bm{H}\right)=\mathcal{N}\left(\bm{z}_{ki} | \boldsymbol{\mu}_{ki}, \operatorname{diag}\left(\boldsymbol{\sigma}_{ki}^{2}\right)\right), \\
    \boldsymbol{\mu}_k = \text{GCN}_{\mu}(\bm{A}', \bm{H}) = \hat{\bm{A}} \bm{H} \bm{W}_{\mu_k}, \\
    \log \boldsymbol{\sigma}_k = \text{GCN}_{\sigma}(\bm{A}', \bm{H}) = \hat{\bm{A}} \bm{H} \bm{W}_{\sigma_k}.
\end{split}
\end{equation}
Here, $\bm{z}_{ki}$ is the $i$th row of $\bm{Z}_k$, and the same for $\boldsymbol{\mu}_{ki}$ and $\log\boldsymbol{\sigma}_{ki}$. Let $\phi_k = \{\bm{W}_{\mu_k}, \bm{W}_{\sigma_k}\}$ , so the encoder can be formulated as $E_k(\bm{A}', \bm{H};\phi_k)$. 
By applying a reparameterization trick, we use three GCN based encoders $E_d(\bm{A}', \bm{H};\phi_d)$, $E_y(\bm{A}', \bm{H};\phi_y)$ and $E_o(\bm{A}', \bm{H}; \phi_o)$ to encode $\bm{H}$ into three latent variables $\bm{Z}_d$, $\bm{Z}_y$ and $\bm{Z}_o$, respectively. 

We also want to claim that our method is not restricted to the specific graph classification task. By replacing GCN with other graph neural networks like GAT \cite{velivckovic2017graph}, our method can be easily extended to different graph learning tasks. 

\subsubsection{Generative Model.}
The generative model in VGAE is given by an inner product among latent variables. Since there are three types of latent variables in our graph decoder, we first employ an MLP layer to aggregate information from all latent variables and improve the expressiveness. Then we apply an inner product between each pair of node representations, which can be formulated as follows: 
\begin{equation}
\begin{aligned}
    & \bm{Z}_g = ReLU([\bm{Z}_d \oplus \bm{Z}_y \oplus \bm{Z}_o]\bm{W}_{g0}) \bm{W}_{g1}, \\
	& p(\bm{A}' | \bm{Z}_g) = \prod_{i=1}^{N} \prod_{j=1}^{N} p\left({a'}_{i j} | \bm{z}_{gi}, \bm{z}_{gj}\right) \\
    \text{with} & \quad p\left(a'_{i j}=1 | \bm{z}_{gi}, \bm{z}_{gj}\right) = \sigma\left({\bm{z}_{gi}}^{\top} \bm{z}_{gj}\right), 
\end{aligned}
\end{equation}
where $\bm{z}_{gi}$ is the $i$th row of $\bm{Z}_g$. Please note that $\oplus$ is the concatenate operation, ${a'}_{ij}$ is the element of $\bm{A}'$ and $\sigma(\cdot)$ is the logistic sigmoid function. Let $\theta_g=\{\bm{W}_{g0}, \bm{W}_{g1}\}$, so the decoder module can be formulated as $D_g(\bm{Z}_d, \bm{Z}_y, \bm{Z}_o;\theta_g)$.

\subsection{Latent Variables Disentanglement}
\subsubsection{Domain Variables Reconstruction.} The disentanglement architecture of our framework is shown in the red dashed box in Figure \ref{fig:model}, and it consists of three modules working together. 
The domain disentanglement module extracts the domain information by training a domain classifier $C_d$. 
The parameters $\theta_{d}$ are learned by minimizing the binary cross-entropy loss $L_d$. So the objective function of the domain disentanglement module is shown as follow:
\begin{equation}
\begin{aligned}
\hat{d_{i}} &= C_{d} (\bm{Z}_d; \theta_{d} ) \\
L(\hat{d_{i}}, d_{i}) &= d_{i} \log (\hat{d_{i}}) + (1-d_{i}) \log (1-\hat{d_{i}})\\
\mathcal{L}_\text{d}\left(\theta_{d}\right) &=
\frac{1}{n_S+n_T} \sum_{g_i \in (G_{S}, G_{T})} L(\hat{d_{i}}, d_{i}),
\end{aligned}
\end{equation}
where $d_i$ is the domain label.

\subsubsection{Semantic Variables Reconstruction.} The $C_y(\bm{z}_y;\theta_y)$ is the label learning module that extracts the semantic information. This is done by training a label classifier $C_y$ on the source domain since labeled data is unavailable in the target domain. As a result, the parameters $\theta_{y}$ in $C_y$ are learned by minimizing the categorical cross-entropy loss $\mathcal{L}_\text{y}$. The objective function of the label disentanglement module is shown as follows:

\begin{equation}
\begin{aligned}
\hat{y_{i}} &= C_{y} (\bm{Z}_y; \theta_{y} ) \\
L(\hat{y_{i}}, y_{i}) &= y_{i} \log (\hat{y_{i}}) + (1-y_{i}) \log (1-\hat{y_{i}})\\
\mathcal{L}_\text{y}\left(\theta_{y}\right) &=
\frac{1}{n_{S}} \sum_{(g_i, y_i) \in G_{S}} L(\hat{y_{i}}, y_{i}),
\end{aligned}
\end{equation}

where $y_i$ is the graph label. Note that we are doing a graph classification task, the read out function is average over the output node features.

\subsubsection{Noise Variables Reconstruction.} In this part, we give a description of noise variables reconstruction model $D_o(\bm{Z}_o;\theta_o)$. In order to disentangle the random latent variables $\bm{Z}_o$, we reconstruct the noise information introduced by data augmentation, i.e., the noise matrix $\bm{A}_o$, which means that we wish our model to infer $\bm{A}_o$ from the augmented data $\bm{A}'$.
The architecture of noise variables reconstructions module $D_o$ is similar to the generative model $D_g$, which can be formulated as follow:
\begin{equation}
\begin{aligned}
& \bm{R} = ReLU(\bm{Z}_o\bm{W}_{n0}) \bm{W}_{n1}, \\
& p(\bm{A}_o | \bm{R}) = \prod_{i=1}^{N} \prod_{j=1}^{N} p\left(a_{ij} | \bm{r}_{i}, \bm{r}_{j}\right) \\
\text { with }& \quad p\left(a_{ij}=1 | \bm{r}_{i}, \bm{r}_{j}\right)=\sigma\left({\bm{r}_{i}}^{\top} \bm{r}_{j}\right),
\end{aligned}
\end{equation}where $\bm{r}_i$ is the $i$th row of $\bm{R}$. Finally, the objective function can be formulated as follows:
\begin{equation}
		\mathcal{L}_\text{o}\left(\theta_{o}\right) = 
		\bm{E}_{q_{\phi_{o}}\left(\bm{Z}_{o} | \bm{A}',\bm{H}\right)}\left[\log P_{\theta_{o}}\left(\bm{A}_o | \bm{Z}_{o} \right)\right]. 
\end{equation}
\subsubsection{Latent Variables Regularization.} In order to exclude the redundant information from the latent variables, we employ an element-wise maximum entropy loss $\mathcal{L}_{e}$ restriction on each latent variables. The above three disentanglement terms ``pull" the relevant information into each latent variables, while the regularizers ``push" the irrelevant information away from the corresponding latent variables. Such regularization ensures the purity of the reconstructed latent variables. The details of the regularizers are shown as follows:
\begin{equation}
\begin{aligned}
 & \mathcal{L}_\text{e} (\phi_{f, d, y, o}) = \frac{1}{n_S+n_T} \sum_{g_i \in (G_{S}, G_{T})} \sum_{k \in (d, y, o)}{L_e\left(\bm{Z}_k\right)} ,\\
 & L_\text{e}(\bm{Z}_k) = \frac{1}{N \times K_k} \sum_{i=0}^{N}{\sum_{j=0}^{K_k}{\sigma(z_{i j}) \log{\sigma(z_{ij})}}},  
\end{aligned}
\end{equation}where $K_k$ is the dimension of type $k$ latent variable. 

\subsection{Model Summary}
We summarize our model as follows. The total loss of our proposed framework can be formulated as:
\begin{equation}
	\label{total_loss}
	\mathcal{L}\left({\phi_{f, d, y, o}}, \theta_{g, d, y, o}\right) =
    \mathcal{L}_{\mathrm{ELBO}}
	+\gamma \mathcal{L}_\text{d} + 
	\alpha \mathcal{L}_\text{y} + \omega \mathcal{L}_\text{o}
	+\delta \mathcal{L}_\text{e},
\end{equation}
where $\gamma, \alpha, \omega $, and $\delta$ are the hyperparameters that control the weight of losses. In our experiment, we let $\gamma=1$, $\alpha=1$, $\omega=0.1$ and $\delta=5$. Small weight for $\mathcal{L}_\text{o}$ can effectively constrain the negative effect brought by dropping and adding edges. And a larger $\delta$ reduces more information from the latent variables. The training of our model can be divided into two steps. In each epoch, we first fix the noise variables reconstruction module (blue blocks in Figure \ref{fig:model}) i.e., $E_o$ and $D_o$, and jointly train the other modules (red blocks in Figure \ref{fig:model}) with clean data on source and target domains. The training algorithm is shown in Algorithm \ref{alg1}. The optimization process is shown as follow:
\begin{equation}
	\left(\hat{\phi_{f}}, \hat{\phi_{d}}, \hat{\phi_{y}}, \hat{\theta_{g}},\hat{\theta_{d}}, \hat{\theta_{y}} \right)= 
	\underset{\phi_{f, d, y}, \theta_{g, d, y}} {\arg \min } 
    \mathcal{L}\left(\phi_{f, d, y, o}, \theta_{g, d, y, o}\right).
    \label{equ:update1}
\end{equation}
In the second step, to prevent model collapse causing by a bad noise matrix, we train the noise variables reconstruction module independently with the augmented data, while the rest of the parameters remain fixed. The optimization process is shown as follow:
\begin{equation}
	\left(\hat{\phi_{f}}, \hat{\phi_{o}}, \hat{\theta_{o}}\right) = 
	\underset{\phi_{f}, \phi_{o}, \theta_{o} }{\arg \min } 
    \mathcal{L}\left(\phi_{f, d, y, o}, \theta_{g, d, y, o}\right).
\end{equation}
All parameters are optimized using the stochastic gradient descent algorithm. In the test phase, we predict the label as follow: 
\begin{equation}
	\hat{y}=C_{y}\left(E_{y}\left(F\left({\bm{A}, \bm{X}};\hat{\phi_{f}}\right) ; \hat{\phi}_{y}\right) ; \hat{\theta_{ y}}\right)
	\label{equ:update2}
\end{equation}
Please note that we input the clean sample without the data augmentation process in the test phase.

Finally, we summarize the pseudo-code of training DGDA as follows:

\begin{algorithm}[htb]
	\renewcommand{\algorithmicrequire}{\textbf{Input:}}
	\renewcommand{\algorithmicensure}{\textbf{Output:}}
	\caption{DGDA training algorithm}
	\label{alg1}
	\begin{algorithmic}[1]
		\REPEAT
        \STATE Randomly select $g_s, y_s$ from $G_S$; 
	    \STATE Randomly select $g_t, y_t$ from $G_T$; 
	    \STATE Forward propagation; 
	    \STATE Update $\phi_{f}, \phi_{d}, \phi_{y}, \theta_{g},\theta_{d}, \theta_{y} $ based on Equation \eqref{equ:update1}; 
	    \STATE Augment $g_s, g_t$ based on Equation \eqref{equ:noise_gen}
	    \STATE Forward propagation; 
	    \STATE Update $\phi_{f}, \phi_{o}, \theta_{o}$ based on Equation \eqref{equ:update2}; 
		\UNTIL Max Iteration or Early Stopping
	\end{algorithmic}  
\end{algorithm}

\begin{table}[htb]
\centering
\caption{Statistics of IMDB\&Reddit.}
    \begin{tabular}{c|ll}
    	\hline
    	Dataset & \textbf{IMDB-Binary} & \textbf{Reddit-Binary} \\
    	\hline
    	\# of Nodes  & 19,773 & 859,254  \\
    	\# of Edges  & 96,531 & 995,508  \\
    	\# of Graphs & 1,000 & 2,000    \\
        Avg Degree   & 4.88   & 1.16     \\
    	\hline
    \end{tabular}
\label{tab:stat1}
\end{table}

\begin{table}[htb]
\centering
\caption{Statistics of Ego-Network.}
    \begin{tabular}{c|llll}
    	\hline
    	Dataset      & \textbf{OAG} & \textbf{Twitter} & \textbf{Weibo} & \textbf{Digg} \\
    	\hline
    	\# of Nodes  & 953,675      & 456,626          & 1,776,950      & 279,630    \\
    	\# of Edges  & 4,151,463    & 12,508,413       & 308,489,739    & 1,548,126  \\
    	\# of Graphs & 499,848      & 499,160          & 779,164        & 244,128    \\
        Avg Degree   & 4.35         & 27.39            & 173.60         & 5.54       \\
    	\hline
    \end{tabular}
\label{tab:stat2}
\end{table}

\section{Experiments}\label{experiment}

In this section, We first describe the experimental setting from Section 4.1 to Section 4.3, 
and then compare DGDA against state-of-the-art baselines in Section 4.4. 
To verify the effectiveness of latent variables regularization and the graph data augmentation in our model, we perform ablation studies in Section 4.5. In Section 4.6, we further give an analysis about the sensitivity of the hyper-parameters. 
In Section 4.7, we describe the implementation details including the model architecture and the hyper-parameter setting. Finally, we analyse the time complexity of our DGDA in Section 4.8.

\begin{table}[htb]
	\centering
	\caption{Node Features of IMDB\&Reddit Dataset.}
    \begin{tabular}{c|l}
		\hline
		Feature Type & Feature Name  \\
		\hline
		\multirow{6}{*}{Vertex}
          & Coreness\cite{batagelj2003m}. \\
        ~ & Pagerank\cite{page1999pagerank}. \\
        ~ & Hub score and authority score\cite{chakrabarti1999mining}. \\
        ~ & Eigenvector Centrality\cite{bonacich1987power}. \\
        ~ & Clustering Coefficient\cite{watts1998collective}. \\
        ~ & Degree\\
        \hline
	\end{tabular}
	\label{tab:ir_features}
\end{table}

\begin{table}[htb]
	\centering
	\caption{Node Features of Ego-network Dataset.}
    \begin{tabular}{c|l}
		\hline
		Feature Type & Feature Name  \\
		\hline
		\multirow{6}{*}{Vertex}
          & Coreness\cite{batagelj2003m}. \\
        ~ & Pagerank\cite{page1999pagerank}. \\
        ~ & Hub score and authority score\cite{chakrabarti1999mining}. \\
        ~ & Eigenvector Centrality\cite{bonacich1987power}. \\
        ~ & Clustering Coefficient\cite{watts1998collective}. \\
        ~ & Rarity (reciprocal of ego user’s degree) \cite{adamic2003friends}. \\
        \hline
        \multirow{1}*{Embedding}
           & 64-dim DeepWalk\cite{perozzi2014deepwalk} embedding.\\
		\hline
        \multirow{5}*{Ego-net}
           & The number/ratio of active neighbors\cite{backstrom2006group}. \\
        ~ & Density of sub-network induced \\
        ~ & by active neighbors\cite{ugander2012structural}. \\
        ~ & Number of Connected components\\
        ~ & formed by active neighbors\cite{ugander2012structural}. \\
		\hline
	\end{tabular}
	\label{tab:ego_features}
\end{table}

\subsection{Datasets}
\subsubsection{IMDB\&Reddit Dataset} It is composed of two datasets: \textbf{IMDB-Binary} and \textbf{Reddit-Binary}. We take each dataset as a domain. The statistics of the IMDB\&Reddit dataset is shown in \ref{tab:stat1}.
\begin{itemize}
    \item \textbf{REDDIT-BINARY}\cite{yanardag2015deep} is a balanced dataset with each graph corresponding to an online discussion thread where each node corresponds to a user \footnote{http://networkrepository.com/REDDIT-BINARY.zip}. An edge is drawn between two nodes if at least one of them responds to another’s comment. This dataset contains top submissions from four popular communities, namely, \textit{IAmA, AskReddit, TrollXChromosomes, atheism}. \textit{IAmA} and \textit{AskReddit} are two question/answer-based communities and \textit{TrollXChromosomes} and \textit{atheism} are two discussion-based communities. The task is to identify whether a given graph belongs to a question/answer-based community or a discussion-based community.
    
    \item \textbf{IMDB-BINARY}\cite{yanardag2015deep} is a movie collaboration dataset \footnote{http://networkrepository.com/IMDB-BINARY.zip}. Each graph corresponds to an ego-network for each actor/actress, where nodes correspond to actors/actresses, and an edge is drawn between two actors/actresses if they appear in the same movie. Each graph is derived from a pre-specified genre (\textit{Romance} or \textit{Action}) of movies, and the task is to classify the genre graph it is derived from. Note that a movie can belong to both genres at the same time. Therefore, [3] discarded movies from \textit{Romance} genre if the movie is already included in the \textit{Action} genre.
\end{itemize}
Note that the original datasets do not contain any node features, so we generate other six different node features based on the graph structure. All used features are shown in Table \ref{tab:ir_features}.

\subsubsection{Ego-network Dataset} This dataset is first used for social influence prediction in \cite{qiu2018deepinf} \footnote{https://github.com/xptree/DeepInf} and we use it for unsupervised graph domain adaptation. It is collected from four different social network platforms: \textbf{OAG}, \textbf{Digg}, \textbf{Twitter} and \textbf{Weibo}. Similarly, we take each social network platform as a domain. The statistics of Ego-network dataset is shown in Table \ref{tab:stat2}. The details of each domain are introduced as follows: 
\begin{itemize}
    \item \textbf{OAG:} OAG (Open Academic Graph) is generated by combining two academic networks: Microsoft Academic Graph and AMiner. The social network is defined to be the co-author network, and the social action is defined to be citation behaviors, i.e., a researcher cites a paper. It is collected from 20 conferences focusing on data mining, information retrieval, machine learning, natural language processing, computer vision, and database. 

    \item \textbf{Digg:} Digg is a news aggregator application that allows users to vote web content, a.k.a, story, up or down. The social network is defined to be the Digg friendship network, and the social action is defined as voting for the story. 
    
    \item \textbf{Twitter:} This dataset was built after monitoring the spreading processes on Twitter before, during, and after the announcement of the discovery of a new particle (Higgs boson) in Jul. 4th, 2012. Similarly, the social network is defined to be the Twitter friendship network, and the social action is defined to be whether a user retweets “Higgs” related tweets.
    
    \item \textbf{Weibo:} Weibo is the most popular microblog application in China. The complete dataset contains the following networks and posting logs of 1,776,950 users between Sep. 28th, 2012 and Oct. 29th, 2012. Social action is defined as reposting (the retweeting behavior in Weibo). 
    
\end{itemize}
We follow the same data prepossess as \cite{qiu2018deepinf}. A fixed-size sub-network is sampled from the ego user's neighborhood via random walking. More concretely, each sample consists of a 50-node graph labeled by the ego user's action status (active or inactive), and the task is to predict the ego users' status. All used features are shown in Table \ref{tab:ego_features}.

\subsection{Baselines}

We not only compare with the baselines of traditional domain adaptation tasks but also compare with the state-of-the-art domain adaptation methods for graph-structured data. Moreover, we also consider the disentanglement-based domain adaptation methods like DIVA \cite{ilse2020diva} and DSR \cite{cai2019learning}. We try three different random seeds on all the methods so we report the mean and variance.

\textbf{Traditional Domain Adaptation Methods}:
\begin{itemize}
    \item Target: A 3-layer GCN that is trained and tested on labeled data from the target domain, and can serve as a
performance upper bound in domain adaptation tasks.
    \item Domain Adversarial Neural Network (\textbf{DANN}) \cite{ganin2016domain}: An adversarial representation learning framework in which one classifier aims to distinguish the domain labels, while the feature extractor coupled with a gradient reverse layer tries to confuse the domain classifier. 
    \item Margin Disparity Discrepancy (\textbf{MDD}) \cite{zhang2019bridging}: A state-of-the-art unsupervised domain adaptation method for computer vision task. 
\end{itemize}

\textbf{Graph Neural Network based Methods}: 
\begin{itemize}
    \item Domain Adaptive Network Embedding (\textbf{DANE)} \cite{zhang2019dane}: It achieves domain adaptive network embedding via shared weight graph convolutional network and adversarial learning regularization. 
    \item Unsupervised Domain Adaptive Graph Convolutional Networks (\textbf{UDA-GCN}) \cite{wu2020unsupervised}: It applies an attention mechanism to integrate global and local consistency and extract cross-domain node embedding with the help of Gradient Reversal Layer\cite{ganin2016domain}. 
\end{itemize}

\textbf{Disentanglement based Methods}: 
\begin{itemize}
  \item Learning Disentangled Semantic Representation for Domain Adaptation (\textbf{DSR}) \cite{cai2019learning}: An unsupervised domain adaptation framework for computer vision task. It disentangles the input image into semantic latent variables and domain latent variables. 
  \item Domain Invariant Variational Autoencoders (\textbf{DIVA}) \cite{ilse2020diva}: A disentanglement based domain generalization method for medical image classification tasks.
\end{itemize}


\subsection{Evaluation Metrics}
We use F1-Score, the harmonic mean of the precision and recall, to quantitatively evaluate all methods, which is also widely used in other related works. It can be formulated as:
\begin{equation}
    F_{1} = 2 \cdot \frac{\text{precision} \cdot \text{recall}}{\text{precision}+\text{recall}}
\end{equation}

\subsection{Experimental Results}

\begin{table*}[h]
\centering
\caption{IMDB\&Reddit dataset test set F1-score(\%).}
\begin{tabular}{c|llllllll}
    \hline
    
    task & Target & DANN & MDD & DANE & UDAGCN & DSR & DIVA & DGDA(Ours)\\
    \hline
    I$\rightarrow$R & ${90.0}_{\pm 1.6}$ & ${74.4}_{\pm 0.6}$ & ${67.3}_{\pm 2.0}$ & ${73.9}_{\pm 1.1}$ & ${73.7}_{\pm 0.9}$ & ${75.7}_{\pm 1.4}$ & ${75.9}_{\pm 0.7}$ & ${\textbf{78.1}}_{\pm 1.0}$ \\
    
    R$\rightarrow$I & ${76.7}_{\pm 2.6}$ & ${73.8}_{\pm 1.4}$ & ${71.0}_{\pm 0.8}$ & ${72.1}_{\pm 1.6}$ & ${73.9}_{\pm 0.7}$ & ${73.2}_{\pm 1.1}$ & ${74.0}_{\pm 1.3}$ & ${\textbf{74.7}}_{\pm 0.9}$ \\
    \hline
\end{tabular}
\label{tab:social}
\end{table*}

\begin{table*}[h]
\centering
\caption{Ego-network datasets test set F1-score(\%). }
\resizebox{\textwidth}{15mm}{
\begin{tabular}{l|llllllllllllll}
		\hline
		Methods &O$\rightarrow$T &O$\rightarrow$W &O$\rightarrow$D &T$\rightarrow$O &T$\rightarrow$W &T$\rightarrow$D &W$\rightarrow$O &W$\rightarrow$T &W$\rightarrow$D &D$\rightarrow$O &D$\rightarrow$T &D$\rightarrow$W &Avg\\

		\hline
        Target & 53.3 & 53.2 & 62.9 & 43.0 & 53.2 & 62.9 & 43.0 & 53.3 & 62.9 & 43.0 & 53.3 & 53.2 & 53.1 \\
        
        DANN   & ${46.7}_{\pm 0.5}$ & ${41.4}_{\pm 0.4}$ & ${42.9}_{\pm 1.4}$ & ${40.2}_{\pm 0.6}$ & ${42.5}_{\pm 0.8}$ & ${41.1}_{\pm 0.4}$ 
               & ${40.2}_{\pm 0.2}$ & ${47.1}_{\pm 0.3}$ & ${41.8}_{\pm 0.6}$ & ${40.2}_{\pm 0.1}$ & ${45.0}_{\pm 0.3}$ & ${41.6}_{\pm 0.2}$
               & 42.6\\
               
        MDD    & ${47.1}_{\pm 0.7}$ & ${40.6}_{\pm 0.5}$ & ${39.8}_{\pm 1.2}$ & ${40.1}_{\pm 0.5}$ & ${43.1}_{\pm 0.8}$ & ${40.6}_{\pm 0.9}$ 
               & ${40.0}_{\pm 0.1}$ & ${47.5}_{\pm 0.2}$ & ${40.7}_{\pm 0.7}$ & ${40.3}_{\pm 0.1}$ & ${45.1}_{\pm 0.2}$ & ${40.2}_{\pm 0.1}$
               & 42.2\\
               
        DANE   & ${42.2}_{\pm 0.6}$ & ${40.1}_{\pm 0.6}$ & ${46.0}_{\pm 1.9}$ & ${40.2}_{\pm 0.2}$ & ${42.3}_{\pm 0.6}$ & ${40.8}_{\pm 1.9}$ 
               & ${40.2}_{\pm 0.2}$ & ${47.7}_{\pm 0.5}$ & ${42.6}_{\pm 1.4}$ & ${39.9}_{\pm 0.1}$ & ${43.3}_{\pm 2.8}$ & ${40.2}_{\pm 0.4}$
               & 42.1\\
               
        UDAGCN & ${40.2}_{\pm 0.6}$ & ${42.1}_{\pm 0.5}$ & ${45.4}_{\pm 1.2}$ & ${40.4}_{\pm 0.2}$ & ${42.9}_{\pm 0.9}$ & ${40.8}_{\pm 0.8}$ 
               & ${40.3}_{\pm 0.1}$ & ${48.3}_{\pm 0.3}$ & ${41.1}_{\pm 0.7}$ & ${40.0}_{\pm 0.6}$ & ${43.4}_{\pm 0.2}$ & ${40.3}_{\pm 0.2}$
               & 42.1\\
               
        DSR    & ${47.7}_{\pm 0.5}$ & ${41.0}_{\pm 0.6}$ & ${43.4}_{\pm 1.8}$ & ${40.2}_{\pm 0.1}$ & ${41.1}_{\pm 0.5}$ & ${41.4}_{\pm 2.0}$ 
               & ${40.3}_{\pm 0.2}$ & ${46.8}_{\pm 0.5}$ & ${43.4}_{\pm 1.4}$ & ${40.2}_{\pm 0.1}$ & ${45.5}_{\pm 2.8}$ & ${41.2}_{\pm 0.5}$
               & 42.7\\
               
        DIVA   & ${47.1}_{\pm 0.5}$ & ${42.0}_{\pm 0.1}$ & ${44.6}_{\pm 0.3}$ & $\textbf{40.7}_{\pm 0.4}$ & ${42.3}_{\pm 0.5}$ & ${45.1}_{\pm 1.0}$ 
               & ${40.8}_{\pm 0.2}$ & ${48.5}_{\pm 0.1}$ & ${43.7}_{\pm 2.2}$ & ${40.9}_{\pm 0.4}$ & ${46.4}_{\pm 0.7}$ & ${41.7}_{\pm 0.1}$
               & 43.6\\
        
        
		\hline
        

		DGDA(Ours) & $\textbf{49.2}_{\pm 0.1}$ & $\textbf{42.9}_{\pm 0.4}$ & $\textbf{49.7}_{\pm 1.4}$ & ${40.3}_{\pm 0.4}$        & $\textbf{44.7}_{\pm 0.6}$ & $\textbf{46.7}_{\pm 1.7}$
		           & $\textbf{41.5}_{\pm 0.3}$ & $\textbf{49.9}_{\pm 0.3}$ & $\textbf{48.3}_{\pm 1.3}$ & $\textbf{41.1}_{\pm 0.4}$ & $\textbf{48.4}_{\pm 0.6}$ & $\textbf{42.7}_{\pm 0.8}$
		           & $\textbf{45.5}$\\ 
        
		\hline

        \end{tabular}}
\label{tab:ego}
\end{table*}

\subsubsection{IMDB\&Reddit Result.} The experiment results of the IMDB\&Reddit dataset are shown in Table \ref{tab:social}. Our method not only outperforms the conventional unsupervised domain adaptation methods but also outperforms the state-of-the-art graph domain adaptation methods on all the domain adaptation tasks. It's remarkable that our method overpasses the comparison methods on task $I \rightarrow R$. However, the improvement of our method on $R \rightarrow I$ is not so remarkable. This is because the dataset size of Reddit-B is larger than that of IMDB-B.
Furthermore, we can find that the performance of our method is very closed to that trained on the target domain (76.7), which is also another reason that leads to the indistinction of our method.

\subsubsection{Ego-network Result.} The experiment results are shown in Table \ref{tab:ego}.
Our method also outperforms all the other baselines on most tasks. We can find that our method achieves significant performance between the analogical domains, e.g., $T\rightarrow W$ and $W \rightarrow T$, which are both micro-blog platforms. What's more, our method also performs well between the domains with large domain discrepancy like $O \rightarrow T$ and $O\rightarrow D$. In some tasks, the improvement is even larger than 4 points. However, the performance of our model on $T \rightarrow O$ is lower than that of UDAGCN and DIVA. 
According to Table \ref{tab:stat2}, domain OAG (O) is sparser than the other domains, so all the compared models perform badly when the target domain is OAG. We also conduct the Wilcoxon signed-rank test on the experiment result with five different random seeds. The p-value is 0.0216, which shows that our method significantly outperforms the baselines.


\subsection{Ablation Study}
To investigate the effectiveness of the latent variables regularization and the random latent variables, we further devise the following three ablated models. 
\begin{itemize}
  \item \textbf{DGDA-D:} We remove the maximum entropy loss on all latent variables.
  \item \textbf{DGDA-O:} We replace the augmented data with the original data.
  \item \textbf{DGDA-M:} We remove the noise variables reconstruction module.
\end{itemize}
The experiment results are shown Table \ref{tab:abl1}.  

\begin{table*}[h]
\centering
\caption{The ablation study results on Ego-network datasets. }
\resizebox{\textwidth}{12mm}{
\begin{tabular}{l|lllllllllllll}
	\hline
	Methods &O$\rightarrow$T &O$\rightarrow$W &O$\rightarrow$D &T$\rightarrow$O &T$\rightarrow$W &T$\rightarrow$D &W$\rightarrow$O &W$\rightarrow$T &W$\rightarrow$D &D$\rightarrow$O &D$\rightarrow$T &D$\rightarrow$W & Avg\\
	
	\hline
    Target & 53.3 & 53.2 & 62.9 & 43.0 & 53.2 & 62.9 & 43.0 & 53.3 & 62.9 & 43.0 & 53.3 & 53.2 & 53.1\\
    
    DGDA-D & ${48.7}_{\pm 0.3}$ & ${42.3}_{\pm 0.6}$ & ${49.0}_{\pm 1.8}$ & ${40.3}_{\pm 0.1}$ & ${42.2}_{\pm 0.3}$ & ${45.1}_{\pm 1.9}$
           & ${41.0}_{\pm 0.4}$ & ${48.3}_{\pm 0.4}$ & ${47.6}_{\pm 0.9}$ & ${41.0}_{\pm 0.4}$ & ${47.7}_{\pm 0.8}$ & ${42.0}_{\pm 0.6}$
           & 44.6 \\
    
    DGDA-O & ${48.6}_{\pm 0.5}$ & ${41.2}_{\pm 0.5}$ & ${46.3}_{\pm 1.1}$ & ${40.2}_{\pm 0.0}$ & ${41.7}_{\pm 0.9}$ & ${42.7}_{\pm 1.5}$
           & ${40.3}_{\pm 0.1}$ & ${49.1}_{\pm 0.3}$ & ${44.5}_{\pm 1.8}$ & ${40.3}_{\pm 0.0}$ & ${48.0}_{\pm 0.7}$ & ${40.5}_{\pm 0.2}$
           & 43.6 \\
           
    DGDA-M & ${48.3}_{\pm 0.1}$ & ${41.0}_{\pm 0.6}$ & ${47.1}_{\pm 2.6}$ & ${40.2}_{\pm 0.0}$ & ${42.1}_{\pm 0.7}$ & ${43.0}_{\pm 2.3}$
           & ${40.4}_{\pm 0.4}$ & ${48.8}_{\pm 0.2}$ & ${44.7}_{\pm 1.6}$ & ${40.2}_{\pm 0.0}$ & ${47.6}_{\pm 0.4}$ & ${41.0}_{\pm 0.6}$
           & 43.7 \\
           
    \hline

	DGDA & $\textbf{49.2}_{\pm 0.1}$ & $\textbf{42.9}_{\pm 0.4}$ & $\textbf{49.7}_{\pm 1.4}$ & $\textbf{40.3}_{\pm 0.4}$ & $\textbf{44.7}_{\pm 0.6}$ & $\textbf{46.7}_{\pm 1.7}$ & $\textbf{41.5}_{\pm 0.3}$ & $\textbf{49.9}_{\pm 0.3}$ & $\textbf{48.3}_{\pm 1.3}$ & $\textbf{41.1}_{\pm 0.4}$ & $\textbf{48.4}_{\pm 0.6}$ & $\textbf{42.7}_{\pm 0.8}$ & \textbf{45.5} \\ 
	\hline

\end{tabular}}
\label{tab:abl1}
\end{table*}

\subsubsection{Study of Latent Variables Regularization.}
To study the effectiveness of the latent variables regularization, we devise \textbf{DGDA-D} in which the regularization loss is removed. The experiment results are shown in Table \ref{tab:abl1}. According to the experiment results, we can find that the ablated model still achieves a comparable performance compared with the state-of-the-art baseline. But the performance drops in all the transfer learning tasks, which shows that the latent variables regularization can restrict these three types of latent variables and contribute to disentanglement.

\subsubsection{Study of Noise Variables Reconstruction.}
To study the effectiveness of manipulation latent variables, we devise \textbf{DGDA-O} and \textbf{DGDA-M}. Comparing with the result of \textbf{DGDA} and \textbf{DGDA-O}, we find that \textbf{DGDA} achieves a better result, which testifies that the data augmentation process is beneficial to learning the noise latent variables. Comparing the result of \textbf{DGDA} and \textbf{DGDA-M}, we also find \textbf{DGDA} performs better than \textbf{DGDA-M}, which testifies that the noise variables reconstruction module improves the model robustness and the transferring performance.

\subsubsection{Study of Graph Data Augmentation on Baselines}
For a fair comparison, we applied the graph data augmentation to the baseline methods. The experiment results on the Ego-networks dataset are shown in Table \ref{tab:abl2}. 

\begin{table*}[h]
\centering
\caption{The ablation study results of graph data augmentation process. All the baselines are trained with both original and augmented data. }
\resizebox{\textwidth}{15mm}{
\begin{tabular}{l|lllllllllllll}
		\hline
		Methods &O$\rightarrow$T &O$\rightarrow$W &O$\rightarrow$D &T$\rightarrow$O &T$\rightarrow$W &T$\rightarrow$D &W$\rightarrow$O &W$\rightarrow$T &W$\rightarrow$D &D$\rightarrow$O &D$\rightarrow$T &D$\rightarrow$W &Avg\\
		
		\hline
        Target & 53.3 & 53.2 & 62.9 & 43.0 & 53.2 & 62.9 & 43.0 & 53.3 & 62.9 & 43.0 & 53.3 & 53.2 & 53.1\\
               
        DANN+Augmented   & ${46.8}_{\pm 0.5}$ & ${41.0}_{\pm 0.2}$ & ${43.3}_{\pm 0.1}$ & ${40.2}_{\pm 0.1}$ & ${42.4}_{\pm 0.3}$ & ${41.6}_{\pm 0.1}$ 
               & ${40.2}_{\pm 0.0}$ & ${47.1}_{\pm 0.3}$ & ${43.3}_{\pm 1.5}$ & ${40.1}_{\pm 0.0}$ & ${46.5}_{\pm 0.3}$ & ${41.6}_{\pm 0.1}$
               & 42.8\\
               
        MDD+Augmented    & ${46.4}_{\pm 0.7}$ & ${42.0}_{\pm 0.1}$ & ${41.8}_{\pm 0.3}$ & ${40.3}_{\pm 0.2}$ & ${43.2}_{\pm 0.4}$ & ${43.0}_{\pm 1.5}$ 
               & ${40.1}_{\pm 0.0}$ & ${47.4}_{\pm 0.3}$ & ${41.3}_{\pm 0.9}$ & ${40.2}_{\pm 0.0}$ & ${43.1}_{\pm 0.6}$ & ${40.7}_{\pm 0.3}$
               & 42.4\\
               
        DANE+Augmented   & ${42.7}_{\pm 0.5}$ & ${40.2}_{\pm 0.6}$ & ${46.1}_{\pm 1.6}$ & ${40.1}_{\pm 0.7}$ & ${41.9}_{\pm 1.7}$ & ${41.4}_{\pm 1.2}$
               & ${40.0}_{\pm 0.1}$ & ${48.0}_{\pm 0.5}$ & ${43.5}_{\pm 1.2}$ & ${40.3}_{\pm 0.2}$ & ${43.5}_{\pm 2.0}$ & ${41.0}_{\pm 0.3}$
               & 42.3\\
               
        UDAGCN+Augmented & ${40.3}_{\pm 0.7}$ & ${42.5}_{\pm 0.8}$ & ${47.0}_{\pm 1.2}$ & ${40.2}_{\pm 0.5}$ & ${42.9}_{\pm 0.5}$ & ${40.6}_{\pm 0.4}$
               & ${40.3}_{\pm 0.1}$ & ${48.2}_{\pm 0.3}$ & ${42.0}_{\pm 0.4}$ & ${40.1}_{\pm 0.1}$ & ${43.1}_{\pm 0.4}$ & ${40.7}_{\pm 0.6}$
               & 42.3\\
               
        DSR+Augmented    & ${47.6}_{\pm 0.1}$ & ${42.2}_{\pm 0.4}$ & ${46.4}_{\pm 1.0}$ & $\textbf{41.1}_{\pm 0.2}$ & ${42.4}_{\pm 0.3}$ & ${43.8}_{\pm 1.2}$ 
               & ${41.3}_{\pm 0.2}$ & ${48.5}_{\pm 0.4}$ & ${46.6}_{\pm 0.4}$ & ${40.4}_{\pm 0.4}$ & ${47.4}_{\pm 0.6}$ & ${42.4}_{\pm 0.1}$
               & 44.2 \\
               
        DIVA+Augmented   & ${47.8}_{\pm 0.5}$ & ${42.0}_{\pm 0.1}$ & ${43.5}_{\pm 0.9}$ & ${40.2}_{\pm 0.3}$ & ${42.5}_{\pm 0.6}$ & ${43.7}_{\pm 1.1}$
               & ${40.8}_{\pm 0.1}$ & ${48.0}_{\pm 0.1}$ & ${45.2}_{\pm 2.2}$ & ${41.0}_{\pm 0.7}$ & ${46.9}_{\pm 0.3}$ & ${41.6}_{\pm 0.1}$
               & 43.6 \\
        

		\hline

		DGDA(Ours) & $\textbf{49.2}_{\pm 0.1}$ & $\textbf{42.9}_{\pm 0.4}$ & $\textbf{49.7}_{\pm 1.4}$ & ${40.3}_{\pm 0.4}$ & $\textbf{44.7}_{\pm 0.6}$ & $\textbf{46.7}_{\pm 1.7}$ & $\textbf{41.5}_{\pm 0.3}$ & $\textbf{49.9}_{\pm 0.3}$ & $\textbf{48.3}_{\pm 1.3}$ & $\textbf{41.1}_{\pm 0.4}$ & $\textbf{48.4}_{\pm 0.6}$ & $\textbf{42.7}_{\pm 0.8}$ & $\textbf{45.5}$\\ 
        
		\hline

        \end{tabular}}

\label{tab:abl2}
\end{table*}
According to the experiment results, most of the compared methods benefit from the augmented training data in different degrees. Especially, DSR gains an average improvement of more than 1 point and achieves the best performance in task $T\rightarrow O$. Even so, our proposed method DGDA still outperforms most of the baselines. This ablation study again validates the superior performance of our method for the graph unsupervised domain adaptation task.

\subsection{Sensitivity Analysis}
\begin{figure*}[htb]
	\centering
	\includegraphics[scale=0.70]{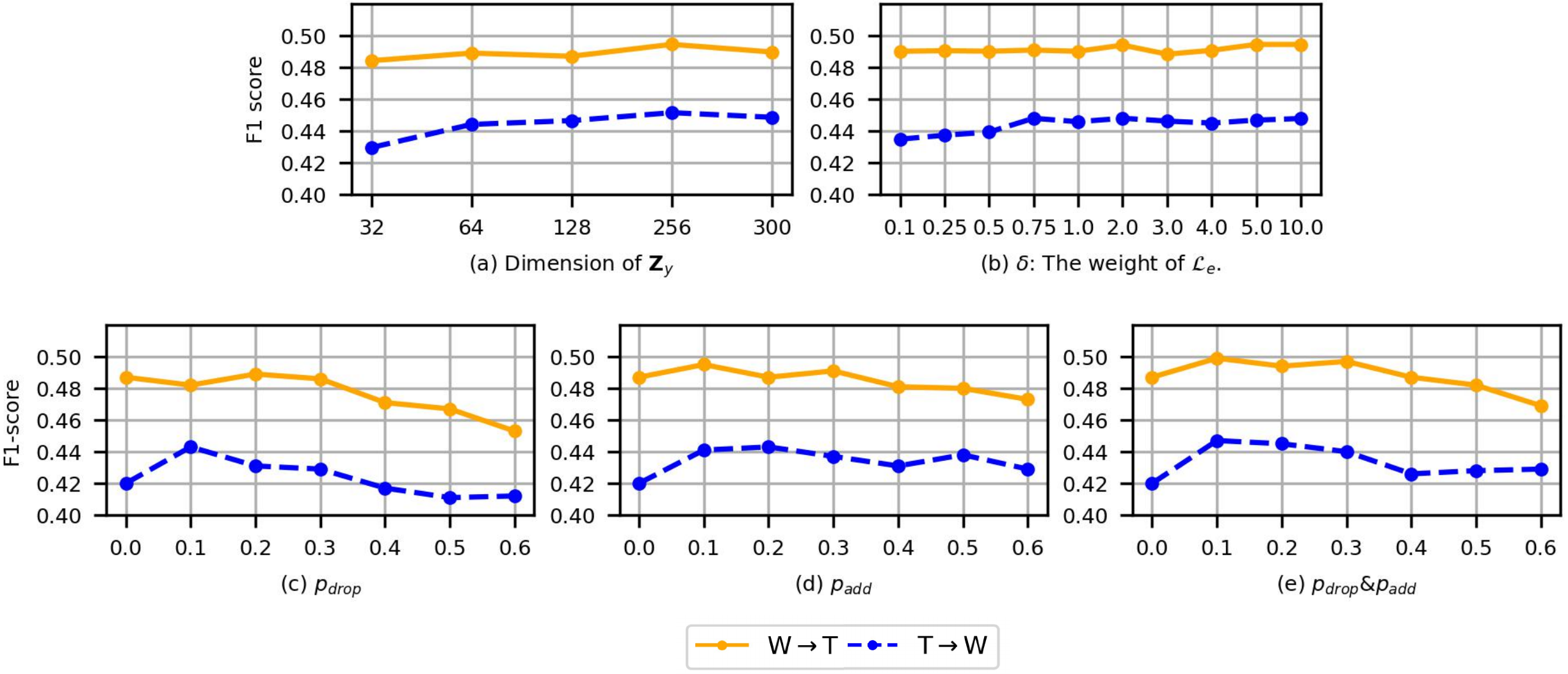}
	\caption{Sensitivities of hyperparameters on the performance of DGDA.}
    \label{fig:hypm}
\end{figure*}

In this section, we investigate the sensitivity of hyperparameters. More specifically, we evaluate how the different dimensions of the semantic latent variables $\bm{Z}_y$, and the latent variables regularization weight $\delta$, as well as the different edge drop rates, affect the performance of \textbf{DGDA}. We report the test set F1 scores on the W$\rightarrow$T and T$\rightarrow$W of the ego-network datasets in Figure \ref{fig:hypm}.

First, As shown in Figure \ref{fig:hypm}(a), we try different dimensions of semantic latent variables $\bm{Z}_y$ from 32 to 300 while fixing the dimension of $\bm{Z}_d$ and $\bm{Z}_o$. We observed that 256-dimension $\bm{Z}_y$ achieves the best performance on both two tasks. Furthermore, when the dimension is larger than 64, only slight differences can be observed with different numbers of dimensions. These results show that DGDA is stable with the dimension of $\bm{Z}_y$.  

Second, as shown in  Figure \ref{fig:hypm}(b) parameter $\delta$ denotes the weight of maximum entropy loss, which controls how much information we exclude from the latent variables. We try different $\delta$ values from 0.1 to 10.0. We find that the performance is stable in both $W \rightarrow T$ and $T \rightarrow W$ when $\delta$ is larger than 1.0. In our experiment, the value of $\delta$ is 5.0.

Finally, Figure \ref{fig:hypm} (c)-(e) show the performance of \textbf{DGDA} in the condition of different edge perturbation ratio. We vary the perturbation rate from 0.0 to 0.6 and find that: (1) the F1 score increases in both tasks as the increment of the edge perturbation ratio and achieves the best result with a 0.1 to 0.3 edge perturbation ratio, which shows that considering the uncertainty of graph data is beneficial to the robustness of the graph neural networks. (2) the F1 score gradually drops when the edge perturbation ratio becomes larger. This is because a too large edge perturbation ratio leads to the destruction of semantic-related structures and results in the degeneration of the performance. As a result, we set $p_\text{drop}=0.1$ and $p_\text{add}=0.1$ in our experiments. 

\begin{figure}[!h]
	\centering
	\includegraphics[scale=0.65]{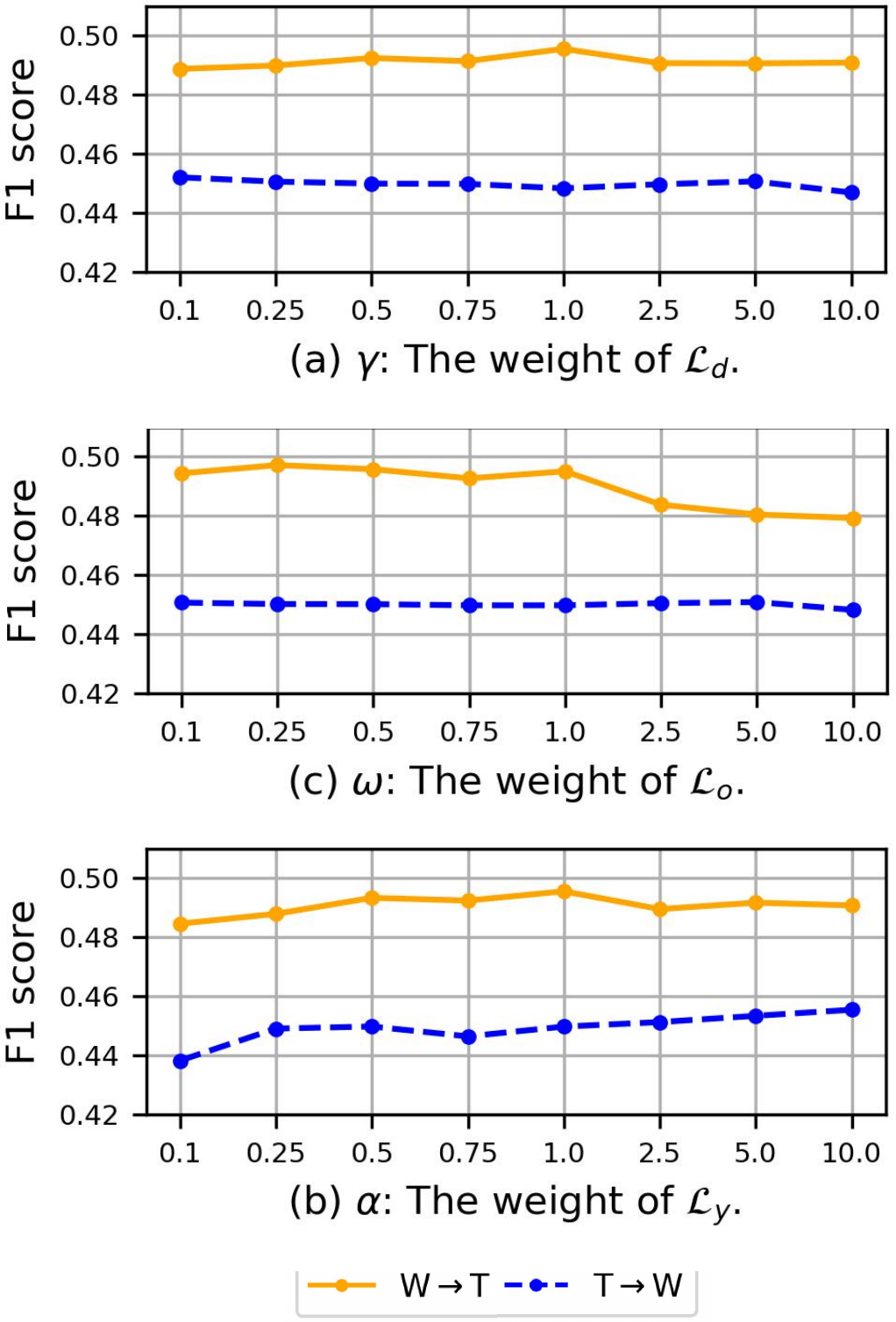}
	\caption{Sensitivities of the loss weights on the performance of DGDA.}
    \label{fig:hypm_dyo}
\end{figure}

We also investigate the sensitivity of the loss weights $\gamma$, $\alpha$, and $\omega$ in Equation (12) of the main manuscript. They correspond to the weight of $\mathcal{L}_\text{d}$, $\mathcal{L}_\text{y}$, and $\mathcal{L}_\text{o}$, respectively. The results are shown in Figure \ref{fig:hypm_dyo}. 

Firstly, as shown in Figure \ref{fig:hypm_dyo} (a), our DGDA is stable with $\gamma$. This is because it is easy to determine the domain label of the graph representation. We set $\gamma=1$ in our experiment.
Secondly, according to Figure \ref{fig:hypm_dyo} (b), we find that the performance drops around 0.01 on both tasks when the value $\alpha$ is small ($\leq$0.25). As $\alpha$ increases, the performance increases and becomes more stable. We set $\alpha=1$ in our experiment.
Thirdly, as shown in Figure \ref{fig:hypm_dyo} (c), we find that the performance is stable when the value of $\omega$ is small ($\leq$1.0) and the performance may drop when $\omega$ becomes larger. This is because a large weight on $\mathcal{L}_\text{o}$ submerges the gradient from $\mathcal{L}_\text{y}$ and causes model collapse. We set $\omega=0.1$ in our experiment.

\subsection{Implementation Details}
For fairness, we use the same architectures in our DGDA and all compared baselines, which is shown in Table \ref{tab:comps}. The other hyper-parameters are introduced in Table \ref{tab:param}. 

\begin{table}[h]{
    \centering
    \caption{The components of our DGDA and all compared baselines. }
    \begin{tabular}{c|c}
		\hline
		\small{Component}  & Description\\
		\hline
		\small{Feature extractor/Encoder}  &  3-layer GCN.  \\ 
		\small{Class classifier}                  & 2-layer MLP. \\
        \small{Domain classifier}             & One linear layer. \\
        \small{Decoder}                             & 2-layer MLP. \\
        \small{Readout function}             & Mean\cite{gilmer2017neural}. \\
        \small{Activation function}         & ReLU\cite{nair2010rectified}. \\
		\hline
	\end{tabular}
	\label{tab:comps}
	
}\end{table}

\begin{table}[h]{
    \centering
    \caption{The hyper-parameters of DGDA. }
    \begin{tabular}{c|cc}
		\hline
		\small{Parameter}  & \textbf{IMDB\&Reddit} & \textbf{Ego-network}\\
		\hline
		\small{Batch size} & 64 & 1024 \\ 
		\small{Learning rate} & 0.001 & 0.01 \\
		\small{Encoder hidden size} & 256 & 256 \\
        \small{Dimension of $\bm{Z}_d$} & 256 & 64 \\
        \small{Dimension of $\bm{Z}_y$} & 256 & 256 \\
        \small{Dimension of $\bm{Z}_o$} & 128 & 128 \\
		\small{Decoder hidden size} & 64 & 64 \\
        \small{Dropout rate} & 0.2 & 0.5 \\
        \small{Weight decay} & 0.0005 & 0.0005 \\
        \small{$p_\text{drop}$} & 0.1 & 0.1 \\
        \small{$p_\text{add}$} & 0.1 & 0.1 \\
		\hline
	\end{tabular}
	\label{tab:param}

}\end{table}

\subsection{Time Complexity Analysis}
In this section, we briefly analyse the time complexity our DGDA. We denote the number of nodes in a graph as $N$, the number of input, hidden, and output dimension as $D_\text{in}$, $D_\text{h}$, and $D_\text{out}$. Assuming that $N > D_\text{in}, D_\text{h}, D_\text{out}$, the computational complexity of one GCN layer is $O(N^3 + N^2 D_\text{out} + N D_\text{in} D_\text{out}) = O(N^3)$, and one linear layer takes $O(N D_\text{in} D_\text{out})$. The time complexity of each block is shown as follows:
\begin{itemize}
  \item Encoder Block: Assuming we employ $L$ layer GCNs, its computational complexity is $O(LN^{3})$.
  \item Decoder Block: it consists of a two-layer MLP and an inner product, so it costs $O(N^{2} D_\text{out} + N D_\text{in} D_\text{h} + N D_\text{h} D_\text{out}) = O(N^{2} D_\text{out})$.
  \item Disentanglement Block: Similarly, it costs $O(N^{2} D_\text{out})$.
\end{itemize}
As a summary, the time complexity of DGDA is $O(L N^3 + N^{2} D_\text{out} + N^{2} D_\text{out})$ = $O(L N^3)$, which is identical to the complexity of stacking $L$ layer GCNs.

\section{Conclusion}\label{conclusion}
In this paper, we present a disentanglement-based framework for the graph unsupervised domain adaptation task. Starting from the generation process of graph-structured data, our approach extracts the disentangled semantic information on the recovered latent space. Specifically, we apply a variational graph auto-encoder architecture with three types of disentanglement modules. Experimental results on two sets of graph classification datasets show that our method outperforms existing methods for graph domain adaptation tasks, which further implies our proposal provides an effective solution for the graph data with high uncertainty and complexity. Following this principle, one promising direction is to go beyond GCN, to investigate more powerful graph neural network architectures for domain adaptation tasks. It would also be interesting to understand and visualize actually what substructure is transferable between graphs from different domains, to provide more interpretability.


%




\ifCLASSOPTIONcaptionsoff
  \newpage
\fi




\bibliographystyle{IEEEtran}
\bibliography{main_2}

%




%
\newpage
\end{document}